\documentclass{article}

\usepackage{microtype}
\usepackage{graphicx}
\graphicspath{{figures/}}
\usepackage{subcaption}
\usepackage{booktabs} 

\usepackage{hyperref}

\usepackage[accepted]{icml2026/icml2026}

\usepackage{amsmath}
\usepackage{amssymb}
\usepackage{mathtools}
\usepackage{amsthm}

\usepackage[capitalize,noabbrev]{cleveref}

\theoremstyle{plain}

\theoremstyle{definition}

\theoremstyle{remark}

\usepackage[disable,textsize=small]{todonotes}
\usepackage[skins,breakable]{tcolorbox}
\usepackage{capt-of}
\usepackage{pgfplots}
\usepackage{pgfplotstable}
\usepackage{xcolor}
\usepackage{url}
\usepackage{comment}

\icmltitlerunning{
On Compositional Learning Behaviours in Formal Mathematics 
}

\begin{document}

\twocolumn[
\icmltitle{
On Compositional Learning Behaviours in Formal Mathematics 
}



  \icmlsetsymbol{equal}{*}

  \begin{icmlauthorlist}
    \icmlauthor{Kevin Yandoka Denamganaï}{yyy}
\end{icmlauthorlist}

  \icmlaffiliation{yyy}{Work conducted as an Independent Researcher during the revision phase of a PhD at the University of York.}

  \icmlcorrespondingauthor{Kevin Yandoka Denamganaï}{kevin.denamganai@york.ac.uk}

  \icmlkeywords{Compositional Learning Behaviours, Compositional Behaviours, Formal Theorem Proving, Lean 4, miniF2F, Symbolic Behaviour, Language Models, In-Context Learning, Chain-of-Thought Reasoning, Compositional Generalisation, Exact Permutation Testing, Self-Evolving Agents, Competence–Performance Distinction, Mathematical Reasoning}

  \vskip 0.3in
]



\printAffiliationsAndNotice{}  

\begin{abstract}
Self-evolving scientific agents capable of conquering the hard tail of formal mathematics require Compositional Learning Behaviours (CLBs)---the capacity to ground and recombine novel symbolic structures in context, beyond mere recombination of prelearned atoms. 
We propose \textbf{S2B-LM}, an adaptation of the CLB-evaluating Symbolic Behaviour Benchmark that removes numerical processing as a confound and adds chain-of-thought scaffolding to elicit rather than merely probe latent CLB competency. 
Cross-evaluating ten Lean~4 theorem provers on CLB competency in S2B-LM and miniF2F whole-proof performance, we find correlational and causal evidence of our claim: 
First, a necessary-condition analysis via quadrant test yields $p=0.004$, with model scale being ruled out as a confound. 
Second, extracting a CLB-encoding activation direction from \textit{DeepSeek-Prover-V2-7B} using S2B-LM traces via Contrastive Activation Addition and applying it during miniF2F whole-proof generation on the AIME subset, CLB suppression collapses solve rate from $32.3\%$ to $2.9\%$, without loss of coherence, while suppressing a random activation direction of equal magnitude leaves it at $31.9\%$.
Together, these results show that CLB competency is \emph{necessary but not sufficient} for the hard tail of formal mathematical verification.
\end{abstract}

\section{Introduction}

\paragraph{Formal Verification and Self-Evolving Agents}
A foundational pillar in the development of self-evolving scientific agents is the reliance on formal verification loops, where automated theorem provers act as execution environments to provide unyielding reward signals for iterative improvement.
Driven by this paradigm, frontier language models have achieved significant performance milestones~\citep{xin2024dspv15,lin2025goedel,ren2025dspv2,lin2025goedelv2} on whole-proof generation tasks by coupling deep neural architectures with intensive proof-tree search algorithms across verification environments such as Lean 4 \citep{zheng2021minif2f, yang2023leandojo, tsoukalas2024putnambench}.

\paragraph{The Diagnostic Gap}
However, standard benchmarking practices typically collapse multidimensional model traits into a monolithic task-accuracy metric. 
This aggregation masks a critical distinction between shallow statistical pattern matching and genuine systematic abstraction. 
True autonomous self-evolution requires the acquisition of Compositional Learning Behaviours (CLBs;~\citet{denamganai2022meta})---the capacity of an agent to ground never-before-seen atomic elements from in-context evidence and recombine them systematically to resolve novel structural configurations within the same episode, going beyond the mere recombination of prelearned atoms (see also~\citet{kim2020cogs}). 
Without explicit diagnostic benchmarks separating across skills, such as CLBs, or structural compliance, being separated from search capacity, it remains ambiguous whether state-of-the-art provers are acquiring authentic symbolic processing capabilities or are merely leveraging expansive compute budgets to overfit to localized tactical distributions that can be seen in formal proving benchmarks.

Following ~\citet{yang2024formal}'s open challenges and future directions, the aim of this paper is to provide a ``controlled stud[y] on synthetic benchmarks for diagnosing reasoning failures'' and highlight a way to ``improve the model architecture for mathematical reasoning''.

We acknowledge the Symbolic Behaviour Benchmark~\citep{denamganai2022meta} as a crucial diagnostic tool to evaluate the capability of state-of-the-art formal theorem proving LMs to perform CLBs. 
However, we note that the original version of the benchmark comes with many confounders when used to evaluate LMs. 
Firstly, the Symbolic Behaviour Benchmark relies heavily on numerical values (cf Section~\ref{sec:method:s2b:domain}), whereas LMs notoriously lack numerical understanding and processing abilities (NUPA - ~\citet{yang2025number}).
Secondly, the Symbolic Behaviour Benchmark does not account for the competence-performance distinction (~\citet{firestone2020performance} - cf.\ Section~\ref{sec:background:human-AI-comparison}), which means that a zero-shot evaluation of CLBs without an adequate reasoning scaffold risks conflating a genuine absence of CLB competency with a mere performance failure---an important confound when interpreting ZSCT scores for LMs.

\paragraph{S2B-LM and Evaluation Protocol}
To address these confounders, we propose \textbf{S2B-LM}, an extension of the S2B that (i) replaces the continuous stimulus domain with a categorical one, removing floating-point segregation as a confound on measured CLB competency, and (ii) introduces a rule-based verbalizer that generates chain-of-thought exemplars from the episode's sync rounds, scaffolding the elicitation of latent CLB competency rather than merely probing it. We report the \emph{adj-ZSCT} (adjusted Zero-Shot Compositional Test score), which maps the raw ZSCT onto an ability-above-chance scale in $[0, 100]$ by removing the intrinsic 50\% random-guessing floor. We cross-evaluate ten state-of-the-art formal theorem proving models, projecting each onto adj-ZSCT alongside a formal verification performance metric. While several benchmarks provide valuable complementary perspectives---including LeanDojo~\citep{yang2023leandojo} and PutnamBench~\citep{tsoukalas2024putnambench}---we focus on miniF2F whole-proof test-split pass@32~\citep{zheng2021minif2f} as it is the evaluation setting we found to be the most consistently reported across papers; notably, 80 of its 244 test-split problems (32.8\%) are drawn from competitive or Olympiad tiers (AMC, AIME, IMO), providing sufficient hard-tail coverage to expose the structural bottleneck we investigate. We then submit the resulting evaluation pool to a necessary-condition quadrant test, and complement the correlational analysis with a causal intervention via Contrastive Activation Addition on S2B-LM traces applied during AIME proof search.

\paragraph{Findings}
We find that CLB competency is a \emph{necessary but not sufficient} condition for elite formal mathematics performance. 
We perform a necessary-condition analysis using a quadrant test, which dichotomises each axis at its empirical median and tests for depletion of a forbidden cell (high miniF2F, low adj-ZSCT) over all $\binom{10}{5}=252$ label placements.
This test finds the forbidden cell \textbf{empty} ($p=0.004$): no model reaches the Olympiad-level tier without measurable CLB competency, while CLB competency alone does not determine ranking within that tier.
Model scale is ruled out as a confound: \textit{DeepSeek-Prover-V2-7B} (7B parameters) clears the Olympiad threshold at $75.6\%$ miniF2F, demonstrating that scale is not a necessary condition for the hard tail.
Beyond this correlational evidence, we seek to establish causality by extracting a CLB-encoding direction from DSP-V2-7B via Contrastive Activation Addition on S2B-LM traces and apply it during miniF2F whole-proof generation on the AIME subset: CLB suppression collapses solve rate from $32.3\%$ to $2.9\%$ while a random direction of equal magnitude leaves it at $31.9\%$, providing direct causal evidence that the CLB representational direction is necessary for Olympiad-tier theorem proving.

\textbf{Contributions.}
\begin{enumerate}
    \item{
    We provide empirical findings that the capability to adapt to novel compositional structures (i.e. receptivity aspects of performing CLBs) is necessary, but not sufficient, for unlocking the hard tail of formal mathematical verification.
    }
    \item{
    We propose the S2B-LM as an extension of the Symbolic Behaviour Benchmark that more precisely evaluates the capabilities of LMs to perform CLBs, by introducing two features to control for distinct confounders: a categorical stimulus domain to remove NUPA as a confound, and a rule-based verbalizer providing chain-of-thought scaffolding to address the competence--performance distinction in LMs.
    }
\end{enumerate}


\section{Background}
\label{background}

\subsection{Compositional Behaviours vs Compositional Learning Behaviours}
\label{sec:background:cb-vs-clb}

\textbf{Definitions.}
\citet{denamganai2022meta} define \emph{Compositional Behaviours} (CBs) as the capacity to generalise from combinations of \emph{trained-on} atomic components to novel re-combinations of those same atoms: the entire vocabulary is prelearned, and generalisation consists of applying known compositional rules to known elements in unseen arrangements. \emph{Compositional Learning Behaviours} (CLBs) extend this to an \emph{online} regime: the agent must generalise from a handful of in-context examples of \emph{never-before-seen} atomic components to novel re-combinations of those atoms, all within a single episode. The within-episode grounding requirement is what distinguishes CLBs from standard compositional generalisation benchmarks, which test training-time generalisation over a prelearned vocabulary.

\textbf{Code generation vs.\ Lean proof generation: CB and CLB in practice.}
Standard code generation is predominantly a CB task: a model trained on Python generalises by composing known identifiers---\texttt{sorted()}, list comprehensions, lambda expressions---into novel arrangements, recombining a fully prelearned vocabulary according to known rules. Lean~4 proof generation introduces a qualitative shift. Lean's reflexive type-checking kernel acts as an unyielding execution environment: every proof step is machine-verified, leaving no room for syntactic hallucination. When a self-evolving prover or a human collaborator introduces a new tactic, a freshly-defined auxiliary lemma, or a novel intermediate proposition---analogous to a mathematician coining a fresh binary relation $\lesssim$ and illustrating its semantics across two or three in-context examples---that object is absent from the model's pretraining weights and cannot be recovered by memorisation. Grounding it from in-context evidence and deploying it correctly in a novel proof context requires decoding the new convention (receptivity) and recombining it with prelearned primitives in unseen arrangements (compositionality)---the full CLB requirement. Thus, while standard code generation and low-to-medium-difficulty theorem proving are accessible via CB capacity alone, advancing into the hard tail of formal mathematical verification demands within-episode symbolic acquisition that current training paradigms do not explicitly cultivate.

\textbf{CLB as symbolic behaviour.}
\citet{santoro2021symbolic} argue that symbolic behaviours do not inhere in agents intrinsically but exist only in relation to an interpreter: a system exhibits symbolic behaviour when another agent treats its outputs as symbols. Among the different aspects of symbolic behaviours, this work focuses on two~\citep{denamganai2022meta}: \emph{receptivity}---the capacity to decode a novel symbolic convention introduced by a speaker agent---and \emph{constructivity}---the capacity to produce a novel convention that a listener agent can decode. The S2B operationalises both by swapping the role assigned to the tested system (cf.\ Section~\ref{sec:background:s2b}); this work focuses exclusively on the receptivity axis. We justify in Section~\ref{sec:receptivity_gatekeeper} why a receptive bottleneck necessarily collapses downstream constructivity, so that receptivity alone is a sufficient diagnostic target.

\subsection{Formal Mathematical Reasoning \& Formal Theorem Proving}
\label{sec:background:formal-math}

Learning-based automated theorem proving in Lean~4 has advanced rapidly: frontier whole-proof generation models---DSP-V1.5~\citep{xin2024dspv15}, Goedel-Prover~\citep{lin2025goedel}, DSP-V2~\citep{ren2025dspv2}, and Goedel-Prover-V2~\citep{lin2025goedelv2}---push miniF2F pass@32 from 48\% to 88\% via SFT on large formal corpora combined with Monte Carlo tree search or RL-based subgoal decomposition. Despite these gains, all operate over a fixed, prelearned tactic vocabulary: they recombine known atomic proof steps (CB), rather than acquiring new symbolic conventions in context (CLB). We evaluate on miniF2F-test whole-proof pass@32~\citep{zheng2021minif2f} (Table~\ref{tab:zsct_minif2f}); while LeanDojo~\citep{yang2023leandojo} and PutnamBench~\citep{tsoukalas2024putnambench} offer complementary perspectives, our choice is guided by the consistent availability of pass@32 results across all ten evaluated models, its single-pass generation regime comparable to the S2B-LM listener evaluation, and its difficulty span---MATH levels 1--5 to IMO, with 80 of 244 test problems from AMC/AIME/IMO---providing the hard-tail coverage needed to expose a CLB bottleneck.

\subsection{How to compare human and AI capabilities ?}
\label{sec:background:human-AI-comparison}

A fundamental challenge in AI evaluation is distinguishing a \emph{lack of capability} from a \emph{failure to demonstrate} it under specific test conditions---the \emph{competence--performance} distinction~\citep{firestone2020performance}, familiar from comparative psychology where subjects cannot be given explicit instructions. Applied to CLB evaluation, an LM may possess the underlying receptive CLB competency yet score near chance if the prompting context does not adequately scaffold its elicitation. Following~\citet{lampinen2024can}, we address this by adapting the prompting context to a few-shot chain-of-thought regime that \emph{elicits} rather than merely probes latent competency (cf.\ Section~\ref{sec:method:s2b:verbalizer}); Section~\ref{sec:method:s2b:validation} provides an ablation study confirming this design choice.

\subsection{Symbolic Behaviour Benchmark}
\label{sec:background:s2b}

\citet{santoro2021symbolic} argue that symbolic behaviours do not reside in agents intrinsically but exist only in relation to an interpreter: a system exhibits symbolic behaviour when another agent treats its outputs as symbols. The Symbolic Behaviour Benchmark (S2B;~\citealt{denamganai2022meta}) operationalises this by requiring a \emph{speaker} agent and a \emph{listener} agent to acquire and deploy a shared symbolic code within \emph{Meta-Referential Games} (Meta-RGs), a meta-learning extension of referential games in which the latent symbolic structure and the communication vocabulary are freshly randomised at every episode.

\textbf{Receptivity and constructivity.}
The S2B evaluates two aspects of CLBs by swapping the role assigned to the tested system. Placing it in the \emph{listener} role---paired with a fixed posdis-compositional rule-based speaker~\citep{chaabouni2020compositionality}---isolates \emph{receptivity}: the capacity to decode an out-of-distribution symbolic convention. Placing it in the \emph{speaker} role---paired with a rule-based listener---isolates \emph{constructivity}: the capacity to produce a novel symbolic convention. In this work we focus exclusively on the receptivity axis.

\textbf{Meta-RG structure and ZSCT.}
Each Meta-RG episode is governed by a latent symbolic structure $(d(1), \ldots, d(N_\text{dim}))$, where $d(i)$ is the number of possible values on latent dimension $i$, and by a freshly sampled random vocabulary permutation that bijectively re-indexes the speaker's tokens before they reach the listener, preventing the listener from exploiting an episode-invariant code. The posdis-speaker encodes latent value $l_i$ at message position $i$ as token $t_i = l_i + 1$ (an offset-1 positional code); the permutation then scrambles these indices, so the listener must rediscover the token-to-value mapping from scratch at every episode. Each game in an episode unfolds over $N$ communication rounds followed by the listener's decision; after each game a dedicated sync round exposes the speaker's exact target stimulus to the listener unconditionally, providing a ground-truth token-to-value update regardless of whether the listener's decision was correct. Episodes unfold in two sequential phases. In the \emph{supporting phase}, games are played until every value on every dimension has appeared in at least $S$ full $N_\text{dim}$-dimensional stimulus vectors (no atomic value is ever presented in isolation). The \emph{querying/ZSCT phase} then presents a held-out set of stimulus combinations---multi-dimensional arrangements of those same values that were withheld as targets during the supporting phase---and accuracy on these games defines the Zero-Shot Compositional Test (ZSCT) metric. Success requires the agent to generalise systematically to novel combinatorial arrangements it has never encountered as targets.

\textbf{Binding problems and the SCS representation.}
Stimuli are encoded using the Symbolic Continuous Stimulus (SCS) scheme: each dimension $i$ partitions $[-1,+1]$ into $d(i)$ sections, and a stimulus entry is a real number sampled from the Gaussian $\mathcal{N}(\mu_{l(i)}, \sigma_{l(i)}^2)$ associated with latent value $l(i)$. The \emph{shape invariance property} ensures that every SCS stimulus is a vector in $[-1,+1]^{N_\text{dim}}$ regardless of the $d(i)$ values, so the structure of the current episode cannot be read off from any single observation: the agent must segregate floating-point coordinates over multiple consecutive supporting-phase games to infer both (i) how many distinct values are active per dimension and (ii) which Gaussian cluster each coordinate belongs to. Sub-problem (i) is the less difficult component of the binding problem; sub-problem (ii) is the harder, numerically demanding component. While this domain-agnostic design is appropriate for RL agents, it introduces a systematic confounder for LMs, whose poor numerical understanding and processing abilities~\citep{yang2025number} directly suppress ZSCT scores independently of any CLB competency.

\section{Method: Adapting the Symbolic Behaviour Benchmark}
\label{sec:method}


\subsection{Adapting the Domain}
\label{sec:method:s2b:domain}

The original S2B represents stimuli using the Symbolic Continuous Stimulus (SCS) scheme~\citep{denamganai2022meta}. Given a latent symbolic structure with $N_\text{dim}$ dimensions described by the tuple $(d(1), \ldots, d(N_\text{dim}))$, a stimulus is a vector in $[-1,+1]^{N_\text{dim}}$: the $i$-th entry is a real number sampled from the Gaussian $\mathcal{N}(\mu_{l(i)},\,\sigma_{l(i)}^2)$ associated with the currently-selected latent value $l(i)\in\{1,\ldots,d(i)\}$ on dimension $i$. Thanks to the shape invariance property (cf.\ Section~\ref{sec:background:s2b}), this scheme instantiates a domain-agnostic binding problem that is appropriate for RL agents. However, for LMs, the Gaussian-sampled real values are a systematic confounder: numerical understanding is not required for compositional reasoning, yet low NUPA~\citep{yang2025number} directly suppresses measured ZSCT scores, breaking the internal validity of the benchmark as a diagnostic of CLBs.

\textbf{The Categorical Domain.}
We replace the SCS representation with a \emph{categorical domain} in which every latent value is a human-readable word. A fixed registry of ten concept classes is defined---\textit{vegetables, fruits, colors, shapes, animals, countries, metals, planets, sports, instruments}---each containing up to ten named items (e.g.\ \textit{colors} $\mapsto$ \{red, blue, green, yellow, \ldots\}). At the start of each episode, $N_\text{dim}$ categories are sampled without replacement from the registry, and for each selected category $i$, exactly $d(i)\in[V_\text{min}, V_\text{max}]$ items are drawn uniformly, forming the episode's latent value set for that dimension. A stimulus on dimension $i$ with latent value $l(i)$ is then presented as the corresponding item name rather than a continuous float. The full $N_\text{dim}$-dimensional stimulus is an ordered tuple of natural language tokens---e.g.\ $(\textit{carrot},\ \textit{blue},\ \textit{circle})$ for a latent symbolic structure with three dimensions using categories \textit{vegetables}, \textit{colors}, \textit{shapes}.

\textbf{Preservation of the Binding Problem.}
The categorical substitution preserves the binding problem structure. The set of active items per dimension is unknown to the tested model \emph{a priori}: although each item name is semantically familiar, the mapping from category items to latent positions is randomised at every episode. The model must therefore integrate evidence across the supporting-phase stimuli to infer which items are active on each dimension and in which combinatorial arrangement---the same latent-structure discovery challenge as in the SCS domain, now expressed entirely in natural language tokens. The train/test combinatorial split strategy and ZSCT evaluation protocol are identical to the original S2B (cf.\ Section~\ref{sec:background:s2b}).

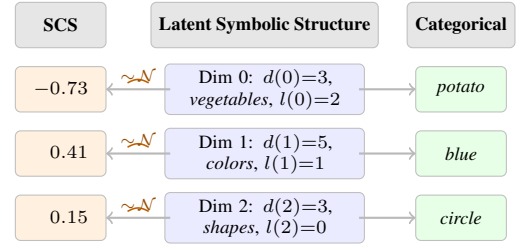
\begin{figure}[t]
  \centering
  \begin{tikzpicture}[
    font=\scriptsize,
    hd/.style={draw=gray!50, fill=gray!20, rounded corners=2pt,
               minimum width=#1, minimum height=0.60cm,
               align=center, font=\scriptsize\bfseries, inner sep=3pt},
    sc/.style={draw=gray!50, fill=orange!12, rounded corners=2pt,
               minimum width=1.2cm, minimum height=0.60cm,
               align=center, font=\ttfamily\scriptsize, inner sep=2pt},
    la/.style={draw=gray!50, fill=blue!8, rounded corners=2pt,
               text width=2.5cm, minimum height=0.60cm,
               align=center, inner sep=2pt},
    ca/.style={draw=gray!50, fill=green!10, rounded corners=2pt,
               minimum width=1.2cm, minimum height=0.60cm,
               align=center, font=\itshape\scriptsize, inner sep=2pt},
  ]
  \def\xs{0}        
  \def\xl{2.7}      
  \def\xc{5.3}      
  \def\dy{-0.85}    

  \node[hd=1.2cm] at (\xs, 0)  {SCS};
  \node[hd=2.85cm] at (\xl, 0)  {Latent Symbolic Structure};
  \node[hd=1.2cm] at (\xc, 0)  {Categorical};

  \node[sc] at (\xs, 1*\dy)  {$-0.73$};
  \node[la] at (\xl, 1*\dy)  {Dim~0: $d(0){=}3$, \textit{vegetables}, $l(0){=}2$};
  \node[ca] at (\xc, 1*\dy)  {potato};

  \node[sc] at (\xs, 2*\dy)  {$\phantom{-}0.41$};
  \node[la] at (\xl, 2*\dy)  {Dim~1: $d(1){=}5$, \textit{colors}, $l(1){=}1$};
  \node[ca] at (\xc, 2*\dy)  {blue};

  \node[sc] at (\xs, 3*\dy)  {$\phantom{-}0.15$};
  \node[la] at (\xl, 3*\dy)  {Dim~2: $d(2){=}3$, \textit{shapes}, $l(2){=}0$};
  \node[ca] at (\xc, 3*\dy)  {circle};

  \foreach \r in {1,2,3} {
    \pgfmathsetmacro\yarrow{\r * \dy}
    \draw[->, thick, gray!55] (\xl-1.25, \yarrow) -- (\xs+0.6, \yarrow);
    \draw[orange!75!black, semithick, domain=-0.14:0.14, samples=18, variable=\t]
      plot ({1.025+\t}, {\yarrow + 0.13*exp(-22*\t*\t)});
    \node[font=\tiny, orange!60!black] at (1.025, \yarrow+0.18) {$\sim\!\mathcal{N}$};
    \draw[->, thick, gray!55] (\xl+1.25, \yarrow) -- (\xc-0.6, \yarrow);
  }
  \end{tikzpicture}
  \caption{SCS vs.\ categorical encoding of the same latent stimulus
    $(l(0){=}2,\,l(1){=}1,\,l(2){=}0)$.
    The \colorbox{orange!12}{SCS scheme} maps each latent value to a
    Gaussian-sampled float ($\sim\!\mathcal{N}$), requiring floating-point
    segregation as a prerequisite to CLB evaluation.
    The \colorbox{green!10}{categorical scheme} replaces floats with named tokens,
    removing that confound while preserving the binding problem.}
  \label{fig:domain_comparison}
\end{figure}

\begin{figure}[t]
  \begin{tcolorbox}[
    enhanced, colback=white, colframe=gray!40,
    boxrule=0.5pt, arc=3pt,
    left=4pt, right=4pt, top=3pt, bottom=3pt,
    fonttitle=\scriptsize\bfseries,
    title={S2B-LM: one supporting-phase game ($N_\text{dim}=3$, \textit{instruments, sports, vegetables})},
  ]
  \scriptsize\setlength{\parskip}{1pt}

  \noindent\textcolor{blue!60!black}{\textbf{[User --- Sync \& Game \#1]}}\\
  Sync: speaker observed \texttt{[piano, swimming, eggplant]} at game~\#0. Your decision was \textit{similar}---\textbf{correct}.\\[2pt]
  At game~\#1, you observe \texttt{[piano, golf, pepper]}. Speaker's message: \texttt{[8,~11,~13]}.\\
  Same latent meaning? Answer~0 (Yes) or~1 (No).

  \medskip\hrule\medskip

  \noindent\textcolor{teal!60!black}{\textbf{[Listener --- Game \#1]}}\\
  Let's think step by step and leverage past games.\\[2pt]
  \textcolor{cyan!60!black}{From the last game syncing, we can learn that: symbol~8 at pos~0 $\to$ \textit{piano}\,;\,symbol~5 at pos~1 $\to$ \textit{swimming}\,;\,symbol~6 at pos~2 $\to$ \textit{eggplant}.}\\[2pt]
  \textcolor{orange!70!black}{In the current game, if the speaker were observing a similar stimulus as ours, [\textit{piano}, \textit{golf}, \textit{pepper}], then: at pos~0, \textit{piano} $\to$ symbol~8 (from game~\#0)\,;\,at pos~1, \textit{golf} has not been observed yet\,;\,at pos~2, \textit{pepper} has not been observed yet.}\\[2pt]
  \textcolor{violet!70!black}{Since the speaker's message is \texttt{[8,~11,~13]}, yield 1/3 matches, they are likely observing a \textit{different} stimulus. \textbf{Answer:~1}}

  \end{tcolorbox}
  \caption{Excerpt of one S2B-LM supporting-phase game illustrating the verbalizer's \textcolor{cyan!60!black}{sync-summary} $\to$ \textcolor{orange!70!black}{inverse-prediction} $\to$ \textcolor{violet!70!black}{match-comparison} reasoning chain. Full conversation in Figure~\ref{fig:conversation_example} (Appendix~\ref{app:conversation}).}
  \label{fig:conversation_excerpt}
\end{figure}

\subsection{Providing Few-Shot Exemplars via a Rule-based Listener Verbalizer}
\label{sec:method:s2b:verbalizer}

A second key design objective of S2B-LM is to address the competence--performance distinction (cf.\ Section~\ref{sec:background:human-AI-comparison}): a tested LM may possess the underlying CLB competency yet score near chance if the prompting context does not adequately scaffold its elicitation. We therefore provide the tested LM with few-shot chain-of-thought exemplars that supply the structured reasoning context needed to \emph{elicit} rather than merely probe its latent CLB competency. We construct these exemplars via a hypothesis-tracking rule-based listener verbalizer, a dedicated agent that monitors the episode and generates human-readable reasoning traces in real time from the sync rounds of the S2B (cf.\ Section~\ref{sec:background:s2b}). The sync round is the game mechanism that unconditionally reveals the speaker's exact target stimulus after every game, making it a natural and game-internal source of ground-truth evidence; the verbalizer converts this evidence into interpretable reasoning steps that are appended to the LM's context before each game decision (see Algorithm~\ref{alg:hypothesis_listener} in Appendix~\ref{app:algorithm} for the full episode loop; Figure~\ref{fig:conversation_excerpt} for a compact illustrative game exchange; and Figure~\ref{fig:conversation_example} in Appendix~\ref{app:conversation} for a complete supporting-phase conversation).

In each supporting-phase game, the tested LM---acting as listener---receives two inputs: its own \emph{stimulus}, an ordered tuple of natural-language category items such as \texttt{[piano, swimming, eggplant]}, and the posdis-speaker's \emph{message}, an integer-token sequence such as \texttt{[8, 5, 6]}. The LM must decide whether both agents are observing stimuli with the same latent meaning. The sync round that follows unconditionally reveals the speaker's exact target stimulus, providing ground-truth evidence on the token-to-value correspondence at every message position regardless of whether the LM's decision was correct.

The rule-based verbalizer maintains a \emph{value map} $\mathcal{V}:\,(\text{pos},\,\text{tok})\mapsto\{\text{val}\mapsto\text{count}\}$ that accumulates this sync-derived evidence across all preceding games in the episode. Before each game, the current state of $\mathcal{V}$ is converted into a natural-language reasoning trace appended to the LM's context, providing chain-of-thought exemplars~\citep{wei2022chain} that prime the LM to replicate the verbalizer's compositional inference on the unseen querying-phase stimuli.

Each verbalized trace follows a three-step template: (i)~\emph{sync summary}---what the most recent sync round revealed about token-to-value correspondences; (ii)~\emph{inverse prediction}---what message the speaker would have sent had it observed the same stimulus as the listener, derived by inverting $\mathcal{V}$; (iii)~\emph{match comparison}---how many positions of the actual message agree with the prediction, yielding the same/different decision. By exposing both the accumulated mapping evidence and the compositional reasoning chain used to reach each answer, these traces prime the tested LM to replicate the same inference pattern on the unseen querying-phase stimuli.

\subsection{Validation via Ablation Study}
\label{sec:method:s2b:validation}

We show via an ablation study that our extended version of the S2B has increased internal validity towards evaluating CLBs in tested LMs.

Because the raw ZSCT operates with an intrinsic 50\% random-guessing floor, we report the \emph{adjusted ZSCT} (adj-ZSCT), which maps ZSCT onto a pure ability-above-chance scale in $[0, 100]$:
\begin{equation}
    \text{adj-ZSCT} = \max\!\left(0,\;\frac{\text{ZSCT} - 50}{100 - 50}\right) \times 100
    \label{eq:adj-zsct}
\end{equation}
Table~\ref{tab:ablation} reports ZSCT and adj-ZSCT scores for DeepSeek-Prover-V2-7B under three benchmark configurations, evaluated over 4 random seeds. Using the smallest model in our suite acts as a conservative lower bound: if the full S2B-LM design successfully elicits CLB competency at 7B, larger models—which exhibit stronger in-context learning capabilities~\citep{brown2020language,wei2022emergent}—can only do better.

\begin{table}[t]
  \centering
  \caption{Ablation results for DSP-V2-7B across benchmark configurations. ZSCT scores reported as mean $\pm$ std.\ err.\ over 4 random seeds. The original S2B uses the SCS domain with no CoT scaffold; S2B-LM uses the categorical domain with 10-shot chain-of-thought exemplars.}
  \label{tab:ablation}
  \resizebox{\columnwidth}{!}{%
  \begin{tabular}{l c c}
    \toprule
    Configuration & ZSCT (\%) & Adj.\ ZSCT \\
    \midrule
    S2B (SCS, 0-shot)         & $57.9 \pm 5.7$ & 15.8 \\
    Categorical + 0-shot      & $40.0 \pm 8.5$ &  0.0 \\
    S2B-LM (Categorical + 10-shot) & $87.0 \pm 4.1$ & 74.0 \\
    \bottomrule
  \end{tabular}%
  }
\end{table}

\begin{table*}[h]
  \centering
  \caption{Evaluation results: adj-ZSCT (S2B-LM receptive listener CLB competency, ability above chance) and miniF2F whole-proof test-split accuracy (Pass@32). ZSCT scores reported as mean $\pm$ std.\ err.\ over 8 random seeds. $\dagger$: model accessed via API.}
  \label{tab:zsct_minif2f}
  \resizebox{\linewidth}{!}{%
  \begin{tabular}{l c l c c c}
    \toprule
    Model & Size & Base Model & ZSCT (\%) & Adj.\ ZSCT ($[0-100]$) & miniF2F test Pass@32 (\%) \\
    \midrule
    DeepSeek-Prover-V1~\citep{xin2024dspv1}              & 7B   & DeepSeekMath-7B~\citep{shao2024deepseekmath}    & $47.6 \pm  6.7$ &   0.0 & 46.1 \\
    DeepSeek-Prover-V1.5-SFT~\citep{xin2024dspv15}      & 7B   & DSP-V1.5-Base~\citep{xin2024dspv15}             & $56.0 \pm  1.6$ &  12.0 & 48.2 \\
    DeepSeek-Prover-V1.5-RL~\citep{xin2024dspv15}       & 7B   & DSP-V1.5-SFT~\citep{xin2024dspv15}              & $56.9 \pm  3.3$ &  13.8 & 50.0 \\
    Goedel-Prover-SFT~\citep{lin2025goedel}              & 7B   & DSP-V1.5-Base~\citep{xin2024dspv15}             & $58.1 \pm  3.0$ &  16.3 & 57.6 \\
    Goedel-Prover-DPO~\citep{lin2025goedel}              & 7B   & Goedel-Prover-SFT~\citep{lin2025goedel}         & $55.3 \pm  3.4$ &  10.6 & 60.3 \\
    Goedel-Prover-V2-8B~\citep{lin2025goedelv2}          & 8B   & Qwen3-8B~\citep{qwen2025qwen3}                  & $75.0 \pm  7.5$ &  50.0 & 84.6 \\
    Goedel-Prover-V2-32B$^\dagger$~\citep{lin2025goedelv2} & 32B  & Qwen3-32B~\citep{qwen2025qwen3}               & $65.2 \pm  6.4$ &  30.4 & 88.1 \\
    DeepSeek-Prover-V2-7B~\citep{ren2025dspv2}           & 7B   & DSP-V1.5-Base-7B~\citep{xin2024dspv15}          & $89.8 \pm  2.7$ &  79.6 & 75.6 \\
    DeepSeek-Prover-V2-671B$^\dagger$~\citep{ren2025dspv2} & 671B & DeepSeek-V3-Base~\citep{deepseek2024v3}        & $100.0 \pm 0.0$ & 100.0 & 82.4 \\
    Kimina-Prover$^\dagger$~\citep{wang2025kimina,wang2025kiminablog} & 72B  & Qwen2.5-72B~\citep{qwen2024qwen25}           & $97.4 \pm  1.6$ &  94.8 & 84.0 \\
    \bottomrule
  \end{tabular}%
  }
\end{table*}

The three configurations isolate the contribution of each design choice. First, the original S2B (SCS, 0-shot) yields only $57.9\%$---marginally above the 50\% guessing floor (adj-ZSCT $= 15.8$)---consistent with NUPA~\citep{yang2025number} suppressing measured CLB scores independently of any compositional competency. Second, switching to the categorical domain while retaining zero-shot prompting (Categorical + 0-shot) drops performance \emph{below} chance ($40.0\%$, adj-ZSCT $= 0$): without a CoT scaffold to elicit the latent reasoning strategy, the model fails to exploit the now-interpretable stimuli, illustrating the competence--performance distinction. Third, the full S2B-LM design (categorical + 10-shot chain-of-thought) recovers strongly to $87.0\%$ (adj-ZSCT $= 74.0$), demonstrating that the verbalizer-generated exemplars successfully surface the model's underlying CLB competency. Together, these results confirm that both design choices---domain substitution and CoT scaffolding---are necessary; neither alone suffices.


\section{Experimental Results}

\textbf{Experimental setup.} miniF2F results report whole-proof test-split accuracy under Pass@32. 
S2B-LM evaluations use $N_\text{dim}=3$ latend dimensions with $V_{\min}=3$ and $V_{\max}=5$ values per latent dimension, number of object-centric samples $O=1$ (the categorical domain does not support object-centric sampling), and number of shots $S=1$ (every atomic element appears in at least one target stimulus before the querying phase begins). 
All reported statistics are computed over $8$ random seeds per model. 
Adj-ZSCT scores are computed via Equation~\ref{eq:adj-zsct}.

Table~\ref{tab:zsct_minif2f} logs the evaluation results across the evaluated model suite. 
Figure~\ref{fig:zsct_vs_minif2f} visualizes the empirical scatter of adj-ZSCT against miniF2F accuracy.

\subsection{Correlational Evidence}
\label{sec:results:correlational}

We perform a necessary-condition analysis via a \emph{quadrant test}: each axis is dichotomised at its empirical median (adj-ZSCT median $= 23.3$; miniF2F median $= 67.95\%$), and we test whether the \emph{forbidden cell}---high miniF2F with low adj-ZSCT---is depleted relative to random pairings. 
Significance is assessed via a one-sided Fisher's exact test by exact enumeration of all $\binom{10}{5}=252$ high-performance label placements (margins fixed). 
Figure~\ref{fig:zsct_vs_minif2f} displays these median thresholds as dashed reference lines.

In the observed data, the forbidden cell is \textbf{empty}: no model achieves high miniF2F with low adj-ZSCT. 
With a resulting p-value of $p=0.004$, this result is establishing CLB competency as a necessary condition for crossing the Olympiad-performance threshold. 
We report on further complementary correlational evidence in Appendix~\ref{app:correlational-evidence}.

\textbf{Limitations.}
Throughout, we rely exclusively on exact permutation methods rather than parametric or asymptotic estimators, as the small evaluation pool ($N=10$) renders standard asymptotic tools structurally invalid. 
We provide further statistical rationale in Appendix~\ref{app:methodology}.
A competing explanation is model scale: larger models may jointly drive both adj-ZSCT and miniF2F performances. 
Crucially, however, \textit{DeepSeek-Prover-V2-7B} (7B parameters) achieves $75.6\%$ miniF2F---placing it in the high-performance half---while remaining a small model. 
This shows that scale is not a necessary condition for the hard tail. 
However, the current result is correlational only. 
Thus, in the next section, we turn to causal evidence to strengthen the CLB account beyond correlation.

\begin{figure}[t]
  \centering
  \begin{tikzpicture}
    \begin{axis}[
      width  = 0.75\columnwidth,
      height = 5.0cm,
      xlabel = {Adj.\ ZSCT: Ability Above Chance (\%)},
      ylabel = {miniF2F Test Accuracy (\%)},
      xmin = -5,  xmax = 105,
      ymin = -5,  ymax = 105,
      xtick = {0,20,...,100},
      ytick = {0,20,...,100},
      axis x line = bottom,
      axis y line = left,
      grid  = major,
      grid style  = {dotted, gray!50},
      tick label style = {font=\small},
      label style      = {font=\small},
    ]


      \addplot[dashed, gray!60, line width=0.8pt] coordinates {(23.3, -5) (23.3, 105)};
      \addplot[dashed, gray!60, line width=0.8pt] coordinates {(-5, 67.95) (105, 67.95)};

      \addplot[
        only marks,
        mark      = *,
        mark size = 2.5pt,
        color     = blue!40!black,
        fill      = blue!70!white,
      ] coordinates {
        (  0.0,  46.1) 
        ( 12.0,  48.2) 
        ( 13.8,  50.0) 
        ( 16.3,  57.6) 
        ( 10.6,  60.3) 
        ( 50.0,  84.6) 
        ( 30.4,  88.1) 
        ( 79.6,  75.6) 
        (100.0,  82.4) 
        ( 94.8,  84.0) 
      };

      \node[anchor=south west, font=\tiny] at (axis cs:  0.0,  46.1) {DP-V1};
      \node[anchor=north west, font=\tiny] at (axis cs: 12.0,  48.2) {DP-V1.5-SFT};
      \node[anchor=south,      font=\tiny] at (axis cs: 13.8,  50.0) {DP-V1.5-RL};
      \node[anchor=west,       font=\tiny] at (axis cs: 16.3,  57.6) {GP-SFT};
      \node[anchor=east,       font=\tiny] at (axis cs: 10.6,  60.3) {GP-DPO};
      \node[anchor=south east, font=\tiny] at (axis cs: 50.0,  84.6) {GP-V2-8B};
      \node[anchor=south,      font=\tiny] at (axis cs: 30.4,  88.1) {GP-V2-32B};
      \node[anchor=north,      font=\tiny] at (axis cs: 79.6,  75.6) {DP-V2-7B};
      \node[anchor=east,       font=\tiny] at (axis cs:100.0,  82.4) {DP-V2-671B};
      \node[anchor=north,      font=\tiny] at (axis cs: 94.8,  84.0) {KP-72B};

    \end{axis}
  \end{tikzpicture}
  \caption{Empirical scatter of adj-ZSCT against miniF2F accuracy. The adj-ZSCT axis captures structured receptive CLB capability adjusted above the 50\% random-guessing baseline.}
  \label{fig:zsct_vs_minif2f}
\end{figure}
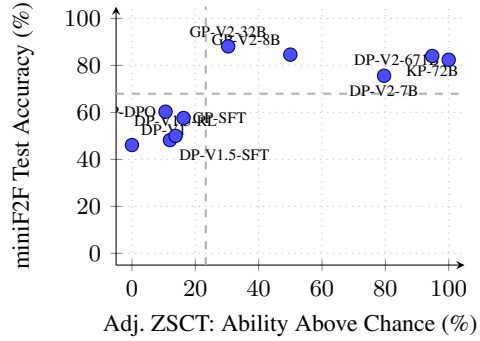

\begin{figure*}[t]
\centering
\tcbset{fontupper=\small\ttfamily, colframe=gray!60, colback=gray!5,
        boxrule=0.5pt, arc=2pt, left=4pt, right=4pt, top=3pt, bottom=3pt}
\begin{minipage}[t]{0.48\textwidth}
  \begin{tcolorbox}[colframe=teal!60!black, title={\small\sffamily\bfseries Correct verbalization (positive)}]
    Let's think step by step and leverage past games.\\[2pt]
    From the last game syncing, we can learn that: \textbf{symbol~3 at pos~0 $\to$ \textcolor{teal!60!black}{violin}} ; \textbf{symbol~5 at pos~0 $\to$ \textcolor{teal!60!black}{trumpet}} ; \textbf{symbol~7 at pos~1 $\to$ \textcolor{teal!60!black}{football}}.\\[2pt]
    In the current game, if the speaker were observing a similar stimulus as ours, [\textbf{violin}, \textbf{football}], then: at pos~0, \textcolor{teal!60!black}{violin} $\to$ symbol~3 (from game~\#2) ; at pos~1, \textcolor{teal!60!black}{football} $\to$ symbol~7 (from game~\#1).\\[2pt]
    Since the speaker's message is [3, 7], yield \textbf{2/2} matches, they are likely observing a \textbf{similar} stimulus.\\[2pt]
    Answer: \textbf{0}
  \end{tcolorbox}
\end{minipage}
\hfill
\begin{minipage}[t]{0.48\textwidth}
  \begin{tcolorbox}[colframe=red!60!black, title={\small\sffamily\bfseries Corrupted verbalization (negative)}]
    Let's think step by step and leverage past games.\\[2pt]
    From the last game syncing, we can learn that: \textbf{symbol~3 at pos~0 $\to$ \textcolor{red!60!black}{trumpet}} ; \textbf{symbol~5 at pos~0 $\to$ \textcolor{red!60!black}{violin}} ; \textbf{symbol~7 at pos~1 $\to$ \textcolor{red!60!black}{football}}.\\[2pt]
    In the current game, if the speaker were observing a similar stimulus as ours, [\textbf{violin}, \textbf{football}], then: at pos~0, \textcolor{red!60!black}{violin} $\to$ symbol~5 (from game~\#1) ; at pos~1, \textcolor{red!60!black}{football} $\to$ symbol~7 (from game~\#1).\\[2pt]
    Since the speaker's message is [3, 7], yield \textbf{1/2} matches, they are likely observing a \textbf{different} stimulus.\\[2pt]
    Answer: \textbf{1}
  \end{tcolorbox}
\end{minipage}
\caption{Synthetic example ($N_\text{dim}=2$) of a matched teacher-forced pair used to extract $\mathbf{v}_{\text{CLB}}$.
Both verbalizations share the identical game structure (message \texttt{[3,\,7]}, listener stimulus \texttt{[violin,\,football]}).
The \textbf{correct} trace (left) uses the true hypothesis map and predicts a 2/2 match, yielding decision~0.
The \textbf{corrupted} trace (right) uses a cyclically shifted binding map at pos~0 (\texttt{symbol~3~$\to$~trumpet}, \texttt{symbol~5~$\to$~violin}) — the text is surface-coherent but the inverse prediction at pos~0 is wrong, reducing to 1/2 matches and flipping the decision to~1.
The per-layer mean residual over each reasoning span is subtracted (correct $-$ corrupted) to obtain the CLB activation direction.}
\label{fig:caa_verbalization}
\end{figure*}

\subsection{Causal Evidence}
\label{sec:results:causal}

\textbf{CLB-encoding direction via CAA.}
To move beyond correlation, we extract a CLB-encoding activation direction from \textit{DeepSeek-Prover-V2-7B} using Contrastive Activation Addition~(CAA), following \citet{turner2023activation} and \citet{rimsky-etal-2024-steering}.
For each querying-phase test game we teacher-force the model with two verbalizations over the \emph{identical} game structure (same symbolic stimulus and speaker message): a \emph{positive} trace produced by the correct rule-based hypothesis tracker (with the true value$\leftrightarrow$symbol binding map accumulated from sync rounds), and a \emph{negative} trace produced by a corrupted tracker whose binding map has been deranged by a cyclic shift---the trace is surface-coherent but reasons from wrong hypotheses, producing wrong inverse predictions and ultimately a wrong listener decision (see Figure~\ref{fig:caa_verbalization} for an illustration).
Because the prompt context is identical across the pair it cancels in the difference. 
Only the CLB reasoning is contrasted between the two traces.
The per-layer mean residual-stream activation over each reasoning span is computed in a matched forward pass, and $\mathbf{v}_{\text{CLB}}$ is the weighted mean of $(\text{correct} - \text{corrupted})$ differences over all matched pairs across episodes.
We record it only over the last quarter of the model's layers (i.e. the $7$ last layers of DeepSeek-Prover-V2-7B).
At inference time, $\alpha \cdot \mathbf{v}_{\text{CLB}}$ is added to the residual stream at each token, where $\alpha > 0$ amplifies and $\alpha < 0$ suppresses the CLB direction.

\begin{figure}[t]
  \centering
  \includegraphics[width=\columnwidth]{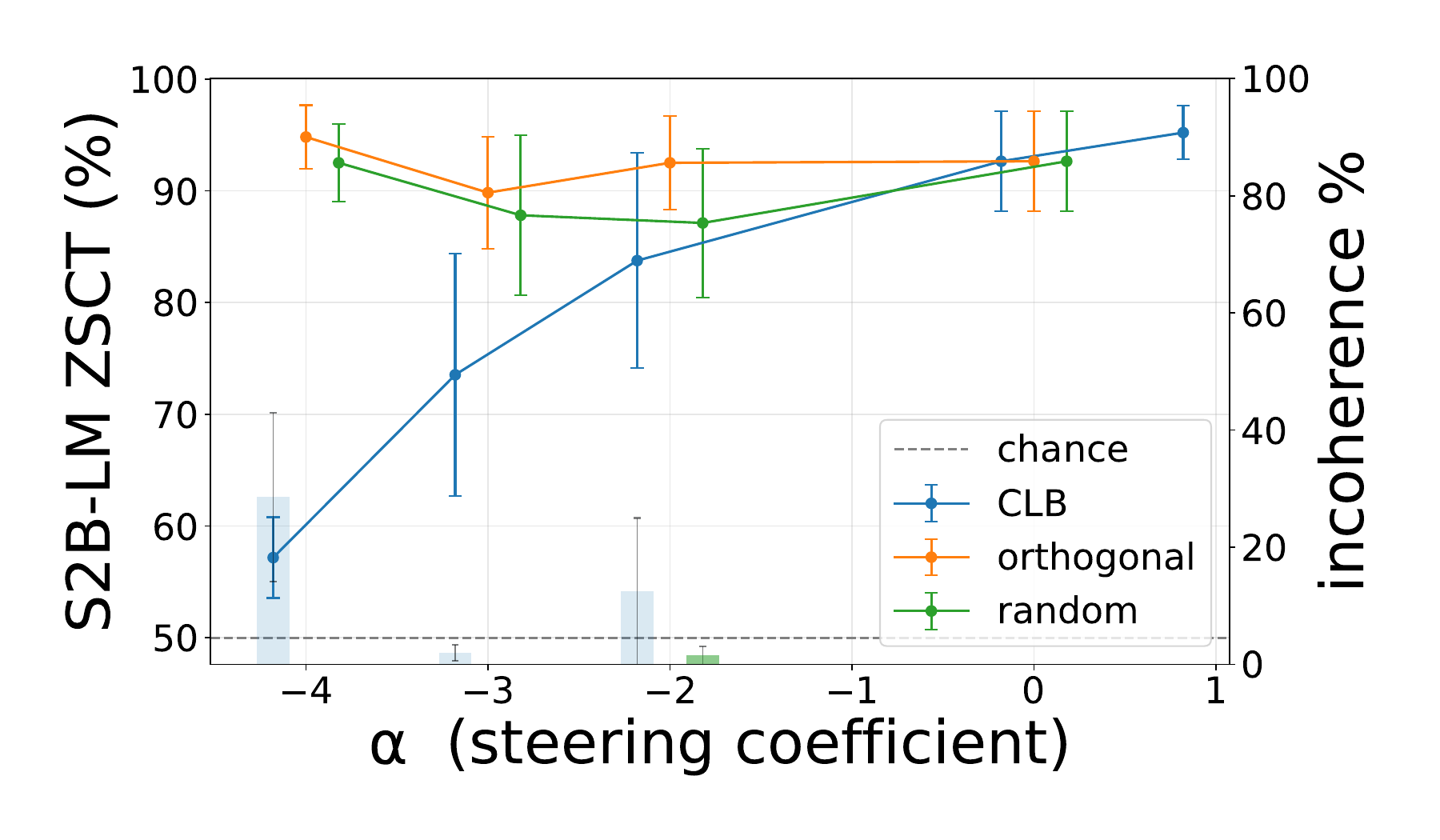}
  \caption{S2B-LM dosage curve for DSP-V2-7B under CLB CAA steering ($\alpha$ on the x-axis), with $n=3$ random seeds per setting.
  Left y-axis: ZSCT (\%); right y-axis: mean output coherence.
  CLB suppression ($\alpha < 0$) monotonically drives ZSCT toward the $50\%$ guessing floor while coherence remains marginally intact, while control directions are flat across the same range.}
  \label{fig:s2b_dosage}
\vspace{-20pt}
\end{figure}

\textbf{Validation on S2B-LM.}
Figure~\ref{fig:s2b_dosage} shows the ZSCT dosage curve as $\alpha$ is swept from $-4$ to $+1$.
CLB steering produces a monotone causal response: suppression at $\alpha = -4$ reduces ZSCT from $92.6\%$ (adj-ZSCT~$= 85.2$) at baseline to $57.2\%$ (adj-ZSCT~$= 14.4$), near the $50\%$ guessing floor, while mild amplification at $\alpha = +1$ yields $95.2\%$ (adj-ZSCT~$= 90.4$).
Two control conditions---an \emph{orthogonal} direction (a random vector projected to be orthogonal to $\mathbf{v}_{\text{CLB}}$) and a \emph{random} direction---show flat responses across the same $\alpha$ range, establishing the specificity of the CLB direction.
We measure \emph{output coherence} as the fraction of generated proofs that are well-formed Lean~4---either proven correct or syntactically valid but unproven (i.e.\ free of parse or elaboration errors)---as opposed to outputs containing syntax errors or empty extractions.
The coherence barplot (Figure~\ref{fig:s2b_dosage}) confirms that CLB suppression does not degrade output coherence: the fraction of well-formed outputs remains stable across all steering magnitudes, ruling out the confound that performance drops stem from garbled generations rather than from suppressed CLB competency.

\begin{figure*}[t]
  \centering
  \includegraphics[width=\textwidth]{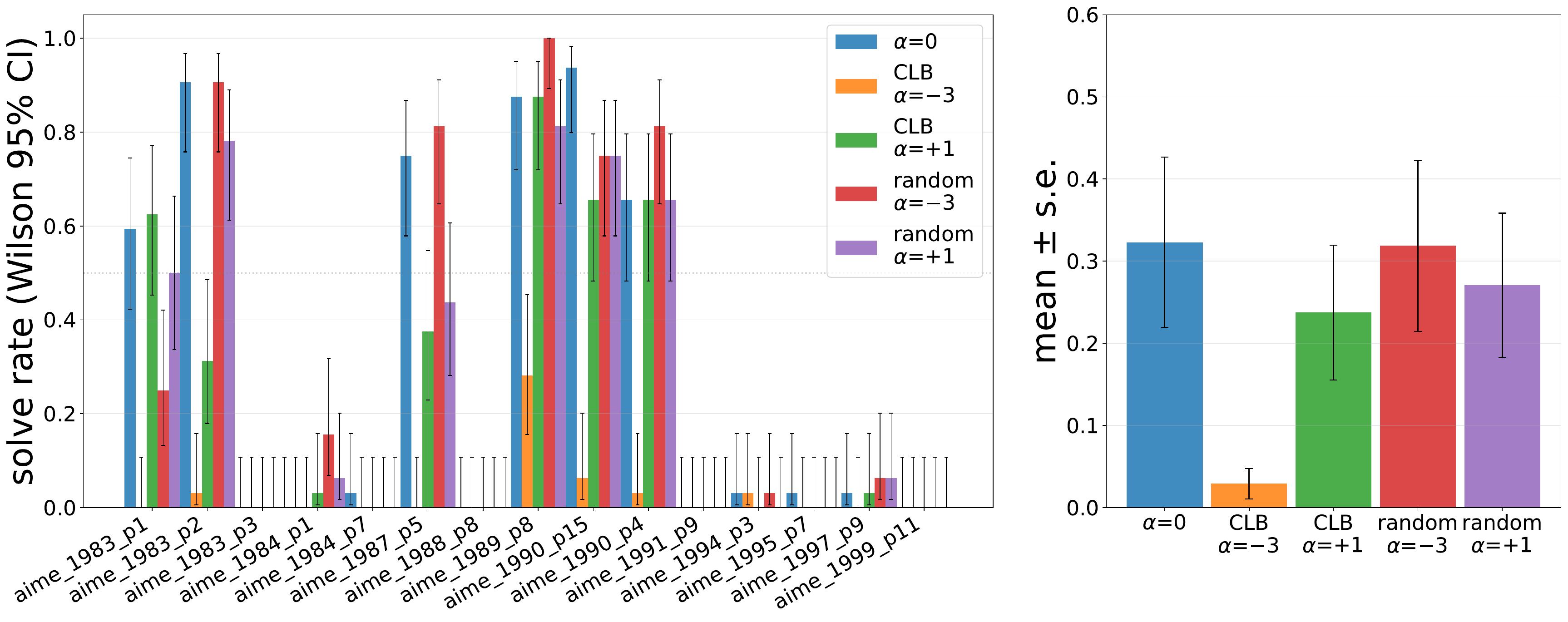}
  \caption{Per-problem (left) and aggregate (right) solve rates (pass@32) under CLB and random CAA steering applied to DSP-V2-7B.
  CLB suppression ($\alpha=-3$) collapses solve rate from $32.3\%$ to $2.9\%$; random suppression leaves it at $31.9\%$.}
  \label{fig:aime_success}
\end{figure*}

\begin{figure}[t]
  \centering
  \includegraphics[width=\columnwidth]{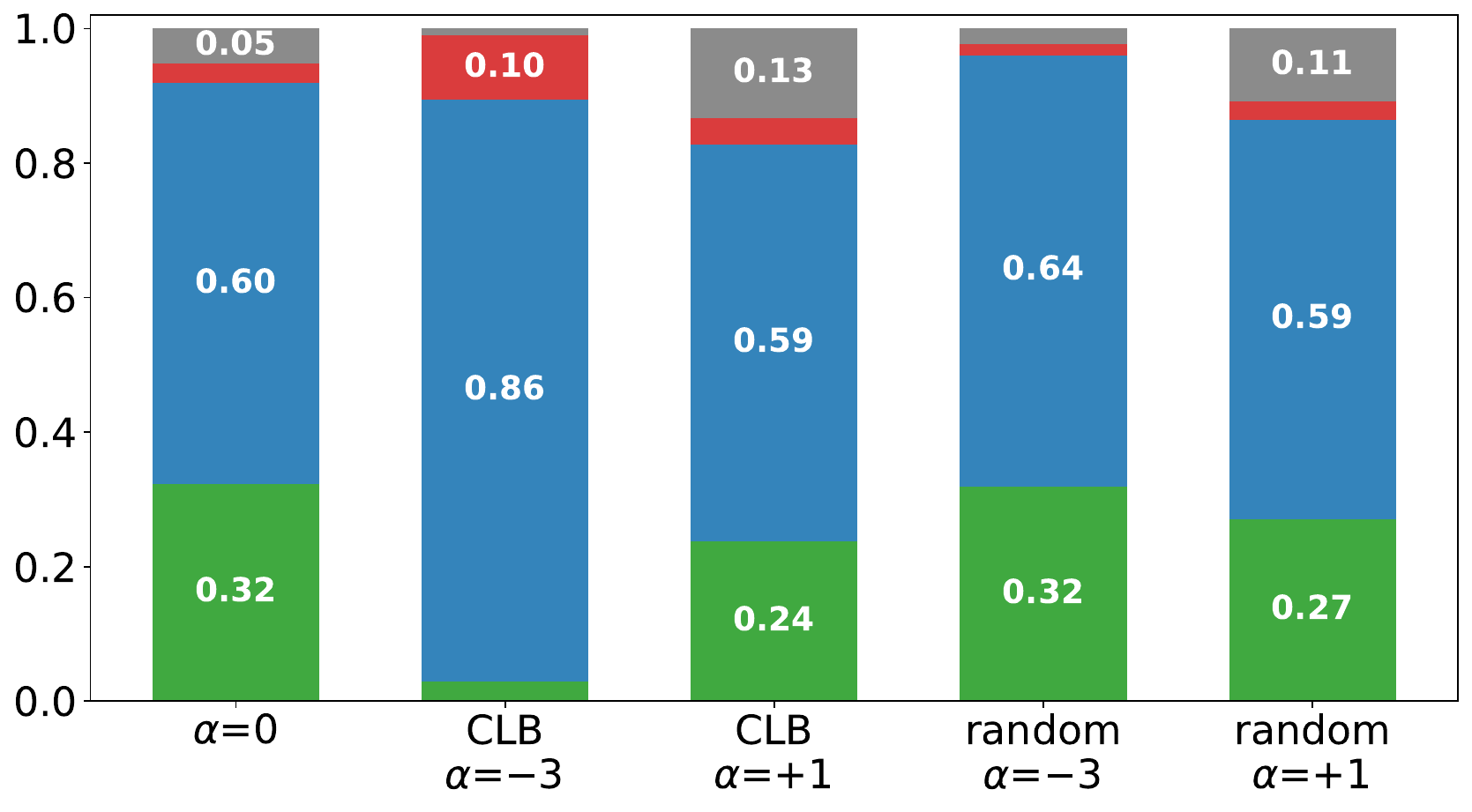}
  \caption{Mean output coherence fractions per condition under CAA steering applied to DSP-V2-7B on the AIME subset.
  Each bar shows the fraction of generated proofs falling into one of four classes: \textcolor[HTML]{2ca02c}{\textbf{solved}} (correct proof), \textcolor[HTML]{1f77b4}{\textbf{valid Lean, unproven}} (well-formed but incomplete), \textcolor[HTML]{d62728}{\textbf{invalid Lean}} (parse/elaboration error), and \textcolor[HTML]{555555}{\textbf{system/timeout}} (Lean compiler timeout or system error).
  Coherence (solved + valid Lean) remains stable across all conditions, ruling out garbled-output confounds.}
  \label{fig:aime_coherence}
\end{figure}

\textbf{Transfer to AIME theorem proving.}
Having validated that $\mathbf{v}_{\text{CLB}}$ selectively modulates CLB competency in S2B-LM, we apply the \emph{same} CLB-encoding CAA vector during miniF2F proof generation, restricting to the $15$ AIME problems in the Olympiad-level, tail subset to maximise the expected CLB signal.
Figures~\ref{fig:aime_success} and~\ref{fig:aime_coherence} show per-problem solve rates (pass@32) and aggregate performance under baseline ($\alpha=0$), CLB suppression ($\alpha=-3$), CLB amplification ($\alpha=+1$), and random controls ($\alpha \in \{-3, +1\}$).

CLB suppression at $\alpha=-3$ collapses aggregate solve rate from $32.3\%$ at baseline to $2.9\%$, without creating incoherent Lean proof candidates. 
This coherent collapse amounts to an $91\%$ relative drop, which is spread nearly uniformly across all 15 problems.
By contrast, the random control at the same magnitude ($\alpha=-3$) leaves performance virtually unchanged at $31.9\%$, confirming that the collapse is specific to the CLB direction and not an artefact of any residual-stream perturbation.
CLB amplification at $\alpha=+1$ yields $23.8\%$---below baseline.
We tentatively attribute this to near-saturation of the model's CLB representations at $\alpha=0$ (ZSCT $92.6\%$, close to ceiling), such that further amplification introduces representational noise rather than additional benefit; however, we acknowledge that this explanation is post-hoc and that the amplification direction may simply be poorly calibrated for proof generation, leaving the asymmetry between suppression and amplification as a limitation for future work.
Together, these results establish a causal role for the CLB direction in AIME-level theorem proving: selectively suppressing it is sufficient to dismantle performance, while the random perturbation of the same norm leaves performance intact.

\section{Discussions}
\label{sec:discussion}

\subsection{On the Asymmetry of Receptivity vs Constructivity in CLBs: Receptivity is a Structural Gatekeeper}
\label{sec:receptivity_gatekeeper}

A potential critique is that a linguistic listener-only benchmark like S2B-LM introduces a task-modality mismatch when evaluating formal theorem provers, which are inherently constructive sequence generators. However, this argument overlooks the structural asymmetry between the two aspects of CLBs established in Section~\ref{sec:background:cb-vs-clb}.
In formal proof synthesis, constructive aspects are structurally downstream of receptive ones.
Even in the whole-proof generation regime, a prover does not synthesise proof steps from a fixed prelearned vocabulary alone: Olympiad-level theorems require the chain-of-thought to decode, bind, and recombine novel intermediate propositions, unfamiliar tactic compositions, and out-of-distribution mathematical objects that cannot be retrieved from pretraining weights by memorisation---the same within-episode grounding requirement that defines the CLB regime (Section~\ref{sec:background:cb-vs-clb}).

If an agent lacks the capacity to systematically process novel, out-of-distribution symbolic structures encountered in context (receptivity), then any subsequent constructive generation degrades.
Confronted with mathematical configurations beyond its compositional horizon, the agent with low receptive CLB competency cannot decompose and recombine the relevant symbolic structures systematically, causing its constructive search to collapse into stochastic matching over its prelearned tactic distribution.

Thus, isolating the receptive axis in S2B-LM is a deliberate diagnostic focus. 
Receptive systematicity serves as a strict structural gatekeeper. 
While low-to-medium difficulty theorems~\citep{zheng2021minif2f} (miniF2F $\le 60.3\%$) can be traversed purely via shallow syntactic matching and possibly massive search budgets without measurable CLB competency, breaking into the hard tail (miniF2F $> 67.2\%$) requires clearing this foundational bottleneck.

\subsection{On Self-Evolving Agents requiring CLBs.}
The stability of a self-evolving scientific agent's improvement loop depends critically on CLB capacity. 
While CB capacity and proof-tree search suffice for the tractable bulk of formal mathematics, a genuinely self-evolving agent must progressively ground proof structures outside its fixed prelearned vocabulary. 
Without robust CLB competency, an agent cannot isolate abstract principles from successful proof trajectories. 
Indeed, its constructive competency would collapse into stochastic matching over prelearned tactics. 
Systematic receptivity thus stabilises the exploration space, enabling the policy to ground newly-encountered symbolic objects and recombine them with prelearned primitives as difficulty increases.

\section{Conclusion}
\label{sec:conclusion}

We investigated the role of Compositional Learning Behaviours (CLBs) in formal mathematical theorem proving by cross-examining ten state-of-the-art provers with the S2B-LM benchmark---an extension of the original S2B that replaces the continuous stimulus domain with a categorical one and introduces chain-of-thought scaffolding, in order to control for NUPA confounders and the competence--performance distinction. Our analysis reveals a fundamental capability dissociation: CLB competency---the receptive ability to ground never-before-seen symbolic objects from in-context evidence and generalise systematically to their novel recombinations, beyond the recombination of prelearned atoms (CB capacity)---is neither uniformly correlated with miniF2F accuracy across the full benchmark, nor orthogonal to it at the Olympiad level.

A necessary-condition analysis via a quadrant test---dichotomising each axis at its empirical median (adj-ZSCT $= 23.3$; miniF2F $= 67.95\%$) and testing depletion of the forbidden cell (high miniF2F, low adj-ZSCT) by exact enumeration of all $\binom{10}{5} = 252$ label placements---finds the forbidden cell \textbf{empty}: no model achieves Olympiad-level performance without measurable CLB competency ($p = 0.004$).
This establishes CLB competency as a necessary, though not sufficient, condition for crossing the Olympiad-performance threshold: CLB competency gates entry to the hard tail without determining ranking within it.
Model scale is ruled out as a confound: \textit{DeepSeek-Prover-V2-7B} (7B parameters) clears the Olympiad threshold at $75.6\%$ miniF2F while remaining a small model, demonstrating that scale is not a necessary condition for the hard tail.

To move beyond correlation, we extract a CLB-encoding direction from DSP-V2-7B via Contrastive Activation Addition on matched teacher-forced S2B-LM pairs---contrasting correct verbalizations against surface-coherent but hypothesis-corrupted ones---and apply it as a residual-stream intervention during miniF2F proof search.
A dosage sweep on S2B-LM confirms that the direction specifically and monotonically modulates CLB competency (ZSCT from $92.6\%$ at $\alpha=0$ to $57.2\%$ at $\alpha=-4$, near the guessing floor) while coherence is preserved, with orthogonal and random control directions producing flat responses.
Transferring the same vector to AIME-level theorem proving, CLB suppression ($\alpha=-3$) collapses the aggregate solve rate from $32.3\%$ to $2.9\%$ ($91\%$ relative drop), while a random direction of equal magnitude leaves performance at $31.9\%$---establishing a specific causal link between the CLB representational direction and Olympiad-tier proof performance.

These findings carry direct implications for self-evolving scientific agents: CB capacity and proof-tree search can navigate the tractable portion of formal mathematics, but hit a combinatorial ceiling without the systematic within-episode generalisation afforded by CLBs. Future work includes: (1) repurposing the S2B-LM to generate synthetic training data that explicitly instils CLB competency into formal proving models; and (2) evaluating the \emph{constructivity} aspects of CLBs---currently unexplored here---by placing tested models in the speaker role of the S2B-LM.

\section*{Impact Statement}
This paper presents work whose goal is to advance the field of Machine Learning. 
There are many potential societal consequences of our work, none which we feel must be specifically highlighted here.

\section*{Acknowledgements}
This work was done as an Independent Researcher, previously supported by the EPSRC Centre for Doctoral Training in Intelligent Games \& Games Intelligence (IGGI) [EP/L015846/1]. 

We gratefully acknowledge the use of Python~\cite{python-2009}, 
Weights and Biases~\citep{wandb}, 
SciPy~\cite{SciPy-NMeth2020}, 
Scikit-learn~\cite{Scikit-learn:JMLR:v12:pedregosa11a}, 
NumPy~\cite{NumPy-Array2020}, 
PyTorch~\cite{pytorch-paszke-NEURIPS2019_9015}, 
HuggingFace's Transformers~\citep{wolf2020huggingfacestransformersstateoftheartnatural}, 
without which this work would not be possible.
We also acknowledge the use of Claude Code (Anthropic) to assist with code development and paper editing throughout this work.\\
The code for this work is available at \url{https://github.com/Near32/OnCLBinFM}, and the S2B-LM codebase is available at \url{https://github.com/Near32/SymbolicBehaviourBenchmark/tree/S2B-LM}.

\bibliography{references}
\bibliographystyle{icml2026/icml2026}

\newpage
\appendix
\onecolumn

\section{Rule-Based Listener Verbalizer Algorithm}
\label{app:algorithm}

\begin{algorithm}[H]
\caption{Hypothesis-tracking rule-based listener verbalizer (one episode)}
\label{alg:hypothesis_listener}
\begin{algorithmic}[1]
\REQUIRE Posdis-speaker $\pi^S$,\; $N_\text{dim}$,\; supporting-phase games $G_\text{sup}$
\STATE $\mathcal{V} \leftarrow \{\}$ \hfill \{value map: $(\text{pos},\text{tok})\to\{\text{val}:\text{count}\}$\}
\STATE $\mathbf{m_\text{last}} \leftarrow \{\}$
\FOR{game $g = 1, \ldots, G_\text{sup}$}
  \STATE Receive speaker message $\mathbf{m}=[t_0,\ldots,t_{N_\text{dim}-1}]$ and own stimulus $\mathbf{s}=[s_0,\ldots,s_{N_\text{dim}-1}]$
  \STATE \textit{// Sync round: update $\mathcal{V}$ from speaker's target and message $m_\text{last}$ at previous game}
  
  \IF{$m_\text{last} \neq \{\}$} 
    \STATE Receive speaker target at previous game, from listener feedback/sync step $\mathbf{f}_\text{last}=[f^\text{last}_0,\ldots,f^\text{last}_{N_\text{dim}-1}]$
    \FOR{$i = 0, \ldots, N_\text{dim}-1$}
        \IF{$t^\text{last}_i \neq \textsc{EoS}$}
            \STATE $\mathcal{V}[(i,\,t^\text{last}_i)][f^\text{last}_i] \mathrel{+}= 1$
        \ENDIF
    \ENDFOR
  \ENDIF
  \STATE \textit{// Bookkeeping: records the current game's speaker message for the next sync round}
  \STATE $\mathbf{m_\text{last}} = [t^\text{last}_0,\ldots,t^\text{last}_{N_\text{dim}-1}] \leftarrow \mathbf{m}$
  \STATE \textit{// Invert $\mathcal{V}$: predict the message the speaker would send for $\mathbf{s}$}
  \FOR{$i = 0, \ldots, N_\text{dim}-1$}
    \IF{$(i,\,s_i)$ has evidence in $\mathcal{V}$ (as a value)}
      \STATE $\hat{t}_i \leftarrow \arg\max_{t}\;\mathcal{V}[(i,\,t)][s_i]$
    \ELSE
      \STATE $\hat{t}_i \leftarrow \textsc{unknown}$
    \ENDIF
  \ENDFOR
  \STATE \textit{// Decision: count matches between actual and predicted message}
  \STATE $n_\text{match} \leftarrow |\{i : \hat{t}_i = t_i,\;\hat{t}_i \neq \textsc{unknown}\}|$
  \STATE $d \leftarrow 0$ (same class) if $n_\text{match} \geq N_\text{dim}$, else $1$ (different class)
  \STATE $\textsc{Emit}\bigl(\textsc{Verbalize}(\mathcal{V},\,\mathbf{s},\,\mathbf{m},\,\hat{\mathbf{t}},\,n_\text{match},\,d)\bigr)$
\ENDFOR
\end{algorithmic}
\end{algorithm}

\section{Conversation Example}
\label{app:conversation}

\begin{tcolorbox}[
  breakable,
  enhanced, colback=white, colframe=gray!40,
  boxrule=0.6pt, arc=3pt,
  left=5pt, right=5pt, top=4pt, bottom=4pt,
  fonttitle=\small\bfseries,
  title={S2B-LM supporting-phase conversation: 6 games, $N_\text{dim}=3$ (\textit{instruments, sports, vegetables}), vocab size 16, max sentence length 3},
]
\scriptsize\setlength{\parskip}{1pt}

\noindent\textcolor{gray!60!black}{\textbf{[System]}}\\
You are a listener agent in a referential game played over multiple rounds. Use the conversation history to learn the speaker's private code across games, then decide whether the current stimulus shares the same latent meaning. Think step by step, then end your response with \textbf{Answer:~0} (same latent meaning) or \textbf{Answer:~1} (different).\\
You and your partner play a sequence of referential games; you are the listener. In the first phase you will get acquainted with the atomic components of the possible observations; then you will be tested with new observations combining the same atomic components in novel ways. At each game each of you observes a stimulus representing a latent meaning, and your common goal is to figure out whether you are observing different or similar latent meanings. Your partner can send you messages via a communication channel of 16 symbols combined into sentences of maximum length~3. Symbol~0 is the end-of-message symbol: any symbol following it is ignored and regularised to~0.

\medskip\hrule\medskip

\noindent\textcolor{blue!60!black}{\textbf{[User --- Game \#0]}}\\
At game \#0, you are observing stimulus: \texttt{[['piano',~'swimming',~'eggplant']]}. Your partner has sent you the following message: \texttt{[8,~5,~6]}.\\
\textit{Question \#1}: At game \#0, do you think you are observing a stimulus with the same latent meaning as your partner? Answer~0: Yes or 1: No. End your response with your decision as a single integer.

\medskip\hrule\medskip

\noindent\textcolor{teal!60!black}{\textbf{[Listener --- Game \#0]}}\\
Let's think step by step and leverage past games. No sync step data yet --- cannot predict expected symbols. \textbf{Answer: 0}

\medskip\hrule\medskip

\noindent\textcolor{blue!60!black}{\textbf{[User --- Sync \& Game \#1]}}\\
At the end of game \#0, sync step: the exact stimulus your partner observed was \texttt{[['piano',~'swimming',~'eggplant']]}. You decided: \textit{similar} latent meanings. This was \textbf{correct} --- you have won game \#0.

At game \#1, you are observing stimulus: \texttt{[['piano',~'golf',~'pepper']]}. Your partner has sent you the following message: \texttt{[8,~11,~13]}.\\
\textit{Question \#1}: At game \#1, do you think you are observing a stimulus with the same latent meaning as your partner? Answer~0: Yes or 1: No. End your response with your decision as a single integer.

\medskip\hrule\medskip

\noindent\textcolor{teal!60!black}{\textbf{[Listener --- Game \#1 (first full verbalized trace)]}}\\
Let's think step by step and leverage past games.
From the last game syncing, we can learn that: symbol~8 at pos~0 $\to$ \textit{piano}\,;\,symbol~5 at pos~1 $\to$ \textit{swimming}\,;\,symbol~6 at pos~2 $\to$ \textit{eggplant}.
In the current game, if the speaker were observing a similar stimulus as ours, [\textit{piano}, \textit{golf}, \textit{pepper}], then: at pos~0, \textit{piano} $\to$ symbol~8 (from game \#0)\,;\,at pos~1, \textit{golf} has not been observed yet\,;\,at pos~2, \textit{pepper} has not been observed yet.
Since the speaker's message is \texttt{[8,~11,~13]}, yield 1/3 matches, they are likely observing a \textit{different} stimulus. \textbf{Answer: 1}

\medskip\hrule\medskip

\noindent\textcolor{blue!60!black}{\textbf{[User --- Sync \& Game \#2]}}\\
At the end of game \#1, sync step: the exact stimulus your partner observed was \texttt{[['piano',~'golf',~'pepper']]}. You decided: \textit{different} latent meanings. This was \textbf{incorrect} --- you have lost game \#1.

At game \#2, you are observing stimulus: \texttt{[['oboe',~'rugby',~'broccoli']]}. Your partner has sent you the following message: \texttt{[3,~9,~2]}.\\
\textit{Question \#1}: At game \#2, do you think you are observing a stimulus with the same latent meaning as your partner? Answer~0: Yes or 1: No. End your response with your decision as a single integer.

\medskip\hrule\medskip

\noindent\textcolor{teal!60!black}{\textbf{[Listener --- Game \#2]}}\\
Let's think step by step and leverage past games.
From the last game syncing, we can learn that: symbol~8 at pos~0 $\to$ \textit{piano}\,;\,symbol~11 at pos~1 $\to$ \textit{golf}\,;\,symbol~13 at pos~2 $\to$ \textit{pepper}.
In the current game, if the speaker were observing a similar stimulus as ours, [\textit{oboe}, \textit{rugby}, \textit{broccoli}], then: at pos~0, \textit{oboe} has not been observed yet\,;\,at pos~1, \textit{rugby} has not been observed yet\,;\,at pos~2, \textit{broccoli} has not been observed yet.
Since the speaker's message is \texttt{[3,~9,~2]}, yield 0/3 matches, they are likely observing a \textit{different} stimulus. \textbf{Answer: 1}

\medskip\hrule\medskip

\noindent\textcolor{blue!60!black}{\textbf{[User --- Sync \& Game \#3]}}\\
At the end of game \#2, sync step: the exact stimulus your partner observed was \texttt{[['oboe',~'rugby',~'broccoli']]}. You decided: \textit{different} latent meanings. This was \textbf{incorrect} --- you have lost game \#2.

At game \#3, you are observing stimulus: \texttt{[['oboe',~'rugby',~'broccoli']]}. Your partner has sent you the following message: \texttt{[3,~15,~6]}.\\
\textit{Question \#1}: At game \#3, do you think you are observing a stimulus with the same latent meaning as your partner? Answer~0: Yes or 1: No. End your response with your decision as a single integer.

\medskip\hrule\medskip

\noindent\textcolor{teal!60!black}{\textbf{[Listener --- Game \#3]}}\\
Let's think step by step and leverage past games.
From the last game syncing, we can learn that: symbol~3 at pos~0 $\to$ \textit{oboe}\,;\,symbol~9 at pos~1 $\to$ \textit{rugby}\,;\,symbol~2 at pos~2 $\to$ \textit{broccoli}.
In the current game, if the speaker were observing a similar stimulus as ours, [\textit{oboe}, \textit{rugby}, \textit{broccoli}], then: at pos~0, \textit{oboe} $\to$ symbol~3 (from game \#2)\,;\,at pos~1, \textit{rugby} $\to$ symbol~9 (from game \#2)\,;\,at pos~2, \textit{broccoli} $\to$ symbol~2 (from game \#2).
Since the speaker's message is \texttt{[3,~15,~6]}, yield 1/3 matches, they are likely observing a \textit{different} stimulus. \textbf{Answer: 1}

\medskip\hrule\medskip

\noindent\textcolor{blue!60!black}{\textbf{[User --- Sync \& Game \#4]}}\\
At the end of game \#3, sync step: the exact stimulus your partner observed was \texttt{[['oboe',~'skiing',~'eggplant']]}. You decided: \textit{different} latent meanings. This was \textbf{correct} --- you have won game \#3.

At game \#4, you are observing stimulus: \texttt{[['drums',~'golf',~'pepper']]}. Your partner has sent you the following message: \texttt{[12,~11,~13]}.\\
\textit{Question \#1}: At game \#4, do you think you are observing a stimulus with the same latent meaning as your partner? Answer~0: Yes or 1: No. End your response with your decision as a single integer.

\medskip\hrule\medskip

\noindent\textcolor{teal!60!black}{\textbf{[Listener --- Game \#4]}}\\
Let's think step by step and leverage past games.
From the last game syncing, we can learn that: symbol~3 at pos~0 $\to$ \textit{oboe}\,;\,symbol~15 at pos~1 $\to$ \textit{skiing}\,;\,symbol~6 at pos~2 $\to$ \textit{eggplant}.
In the current game, if the speaker were observing a similar stimulus as ours, [\textit{drums}, \textit{golf}, \textit{pepper}], then: at pos~0, \textit{drums} has not been observed yet\,;\,at pos~1, \textit{golf} $\to$ symbol~11 (from game \#1)\,;\,at pos~2, \textit{pepper} $\to$ symbol~13 (from game \#1).
Since the speaker's message is \texttt{[12,~11,~13]}, yield 2/3 matches, they are likely observing a \textit{different} stimulus. \textbf{Answer: 1}

\medskip\hrule\medskip

\noindent\textcolor{blue!60!black}{\textbf{[User --- Sync \& Game \#5]}}\\
At the end of game \#4, sync step: the exact stimulus your partner observed was \texttt{[['drums',~'golf',~'pepper']]}. You decided: \textit{different} latent meanings. This was \textbf{incorrect} --- you have lost game \#4.

At game \#5, you are observing stimulus: \texttt{[['guitar',~'rugby',~'eggplant']]}. Your partner has sent you the following message: \texttt{[4,~9,~6]}.\\
\textit{Question \#1}: At game \#5, do you think you are observing a stimulus with the same latent meaning as your partner? Answer~0: Yes or 1: No. End your response with your decision as a single integer.

\medskip\hrule\medskip

\noindent\textcolor{teal!60!black}{\textbf{[Listener --- Game \#5]}}\\
Let's think step by step and leverage past games.
From the last game syncing, we can learn that: symbol~12 at pos~0 $\to$ \textit{drums}\,;\,symbol~11 at pos~1 $\to$ \textit{golf}\,;\,symbol~13 at pos~2 $\to$ \textit{pepper}.
In the current game, if the speaker were observing a similar stimulus as ours, [\textit{guitar}, \textit{rugby}, \textit{eggplant}], then: at pos~0, \textit{guitar} has not been observed yet\,;\,at pos~1, \textit{rugby} $\to$ symbol~9 (from game \#2)\,;\,at pos~2, \textit{eggplant} $\to$ symbol~6 (from game \#3).
Since the speaker's message is \texttt{[4,~9,~6]}, yield 2/3 matches, they are likely observing a \textit{different} stimulus. \textbf{Answer: 1}
\end{tcolorbox}
\captionof{figure}{S2B-LM conversation examples over 6 supporting-phase games illustrating the hypothesis-tracking rule-based listener verbalizer. \textcolor{gray!60!black}{\textbf{System}} sets the task context. \textcolor{blue!60!black}{\textbf{User}} messages present each game's stimulus and message, followed by the sync-step revelation of the speaker's target. \textcolor{teal!60!black}{\textbf{Listener}} responses are produced by $\textsc{Verbalize}$ (Algorithm~\ref{alg:hypothesis_listener}): game~\#0 defaults to \textit{same} (no sync data yet); subsequent games accumulate sync-derived token-to-value evidence in $\mathcal{V}$ and apply the sync-summary $\to$ inverse-prediction $\to$ match-comparison chain; games~\#3--\#5 illustrate how the evidence base grows across episodes, with partial matches (1/3, 2/3) correctly classified as \textit{different} even as the mapping fills in.}
\label{fig:conversation_example}

\section{Further Correlational Evidence}
\label{app:correlational-evidence}

We cross-evaluate the ten provers of Table~\ref{tab:zsct_minif2f} against two complementary non-parametric tests to determine whether CLB competency—as measured by adj-ZSCT—constitutes a necessary prerequisite for high miniF2F performance. 
We begin with a global ceiling-zone analysis over all $10! = 3{,}628{,}800$ pairings (Section~\ref{sec:results:global}). 
Seeing that this global test falls just short of conventional significance ($p = 0.052$), we consider the likely confound introduced by the heavily right-skewed difficulty distribution of miniF2F~\citep{zheng2021minif2f}: search-tractable theorems in the lower range can be solved without any measurable CLB competency, diluting any global boundary signal. 
We therefore apply a structurally targeted tail partition test (Section~\ref{sec:results:tail}) that isolates the Olympiad-level regime (miniF2F $> 75\%$) and asks whether the models that penetrate it cluster exclusively at high adj-ZSCT. 
To rule out model scale as a competing explanation, we embed both predictors in an identical permutation space and compare their significance footprints (Section~\ref{sec:results:scale}). 
Throughout, we rely exclusively on exact permutation methods rather than parametric or asymptotic estimators, as the small evaluation pool ($N=10$) renders standard asymptotic tools structurally invalid; the full statistical rationale is given in Appendix~\ref{app:methodology}.


\subsection{Global Continuous Bottleneck Test}
\label{sec:results:global}

We first apply a global, continuous bottleneck check using non-parametric ceiling zone analysis \citep{dul2016necessary}. This framework operationalises the prerequisite hypothesis geometrically: under a true bottleneck, the upper-left region of the (adj-ZSCT, miniF2F) scatter—high miniF2F, low adj-ZSCT—should be systematically vacant.

\textbf{Limitation: slope dependency.}
The vacancy penalty statistic requires a boundary line $Y = s \cdot X$ that separates the forbidden upper-left region from the permitted lower-right region. Because adj-ZSCT and miniF2F are expressed on the same $[0,100]$ percentage scale but measure incommensurate quantities, the slope $s$ cannot be derived from first principles — it is a free parameter. We set it in a data-determined, scale-free way as $s = \max(\text{miniF2F}) / \max(\text{adj-ZSCT}) = 88.1 / 100.0 = 0.881$, which normalises each axis by its observed maximum so that the boundary passes through the top-right corner of the empirical scatter. This is the most principled choice available, but the p-value remains sensitive to it (e.g.\ $p = 0.050$ at $s = 0.88$, rising to $p = 0.054$ at the bare diagonal $s = 1.0$). The slope-free quadrant test in Section~\ref{sec:results:correlational} avoids this limitation entirely.

\textbf{Null Hypothesis ($\mathcal{H}_0^{\text{CLB,global}}$).} The joint distribution of (adj-ZSCT, miniF2F) pairs exhibits no directional vacancy in the upper-left quadrant; the observed triangular layout is consistent with a random re-pairing of the empirical adj-ZSCT and miniF2F marginal distributions.

\textbf{Alternative Hypothesis ($\mathcal{H}_1^{\text{CLB,global}}$).} The upper-left quadrant of the (adj-ZSCT, miniF2F) space is systematically empty, indicating that low CLB competency constrains the maximum achievable formal verification accuracy.

We define the test statistic $T$ as the negative upper-left vacancy penalty:
\begin{equation}
    T = -\sum_{i=1}^{N} \max\!\left(0,\; \frac{\text{miniF2F}_i}{100} - s \cdot \frac{\text{adj-ZSCT}_i}{100}\right)
\end{equation}
$T = 0$ indicates no violations of the prerequisite ordering; increasingly negative values reflect models that achieve high miniF2F despite low adj-ZSCT. Under $\mathcal{H}_1^{\text{CLB,global}}$, the observed pairing should yield a $T$ value significantly closer to zero than under random re-pairings.

An exact pairings permutation test evaluating all $10! = 3{,}628{,}800$ possible (adj-ZSCT, miniF2F) pairings, using the scale-normalised slope $s = 0.881$, yields an observed statistic of $T = -3.237$ and an empirical $p_{\text{CLB}}^{\text{global}} = \mathbf{0.050}$. This sits exactly at the conventional significance threshold ($\alpha = 0.05$) and should be interpreted with caution given its sensitivity to the slope choice noted above. The primary driver of the residual penalty is \textit{Goedel-Prover-V2-32B}, which achieves miniF2F $= 88.1\%$ at a comparatively low adj-ZSCT $= 30.4$, placing a data point in the upper-left region and inflating the observed vacancy penalty. We attribute this anomaly to the competence--performance distinction (Section~\ref{sec:method:s2b:verbalizer}): the model may possess latent CLB competency that the 10-shot scaffold fails to elicit at this scale and model family, causing adj-ZSCT to underestimate its true CLB ability.

\subsection{Tail Partition Permutation Test}
\label{sec:results:tail}

The failure of the global continuous test motivates a more targeted, structurally motivated hypothesis. The difficulty distribution of miniF2F is heavily right-skewed~\citep{zheng2021minif2f}: 164 of the 244 test-split problems (67.2\%) are drawn from MATH levels 1--5 and custom sources, forming a low-to-medium difficulty bulk solvable via syntactic fluency and extensive proof-tree search without requiring systematic compositional generalisation, while the remaining 80 problems (32.8\%) are drawn from AMC, AIME, and IMO---the competitive and Olympiad tiers that resist brute-force enumeration and should expose a genuine CLB bottleneck. Any global boundary test will therefore be confounded by the abundance of search-solvable theorems in the lower miniF2F range.

We formalise this observation as a \emph{tail partition test} anchored at the structural undergraduate--Olympiad boundary of $67.2\%$: 164 of 244 test-split problems (67.2\%) are MATH-level bulk; the remaining 80 (32.8\%) are AMC/AIME/IMO Olympiad problems. The empirical median of miniF2F across our ten models ($67.95\%$) corroborates this structural cut. The five models that surpass this threshold—\textit{Goedel-Prover-V2-8B} ($84.6\%$), \textit{Goedel-Prover-V2-32B} ($88.1\%$), \textit{DeepSeek-Prover-V2-7B} ($75.6\%$), \textit{DeepSeek-Prover-V2-671B} ($82.4\%$), and \textit{Kimina-Prover} ($84.0\%$)—constitute the \emph{hard tail} partition $\mathcal{T}$.

\textbf{Null Hypothesis ($H_0^{\text{tail}}$).} The five models with the highest miniF2F scores are drawn uniformly at random from the full pool with respect to their adj-ZSCT scores; any apparent clustering at high adj-ZSCT is attributable to chance.

\textbf{Alternative Hypothesis ($H_1^{\text{tail}}$).} The five models that penetrate the Olympiad-level tail exhibit systematically higher CLB competency (adj-ZSCT) than would be expected under a random partition of the pool.

The test statistic is the aggregate adj-ZSCT sum over the hard tail:
\begin{equation}
    S_{\mathcal{T}} = \sum_{i \in \mathcal{T}} \text{adj-ZSCT}_i
\end{equation}
We enumerate all $\binom{10}{5} = 252$ possible ways to select five models from the pool and compute $S_{\mathcal{T}}$ for each partition. The observed value is $S_{\mathcal{T}} = 354.80$. Strikingly, this value corresponds to the \emph{maximum achievable} adj-ZSCT sum across all 252 partitions: the five models in $\mathcal{T}$ are identically the five models with the highest adj-ZSCT scores in the full pool. Exactly \textbf{1} configuration out of 252 meets or exceeds this value, yielding an exact empirical $p$-value of $p_{\text{CLB}}^{\text{tail}} = \frac{1}{252} = \mathbf{0.00397}$. 
This result is highly statistically significant ($p < 0.01$), supporting $H_1^{\text{tail}}$. The set of models that conquer the Olympiad-level tail of formal mathematics is in perfect correspondence with the set of models that exhibit the strongest CLB competency—a coincidence with probability less than $0.4\%$ under the null. Crucially, this finding is consistent with CLB competency being a \emph{necessary} condition for elite formal verification performance: no model outside the top adj-ZSCT tier enters the hard tail, regardless of architecture scale or search budget.

\begin{tcolorbox}[
  enhanced,
  colback=blue!6, colframe=blue!30, boxrule=0.6pt, arc=3pt,
  left=6pt, right=6pt, top=4pt, bottom=4pt,
]
\textbf{Takeaway 1.} CLB competency is not required for the tractable bulk ($p_{\text{CLB}}^{\text{global}} = 0.052$, non-significant) but is a strict structural prerequisite for the Olympiad-level tier: the top-5 models by miniF2F are precisely the top-5 by adj-ZSCT across all $\binom{10}{5}=252$ partitions ($p_{\text{CLB}}^{\text{tail}} = 0.004$).
\end{tcolorbox}

\subsection{Decoupling CLB Competency from Model Scale}
\label{sec:results:scale}

A compelling competing explanation for the observed capability dissociation is the \emph{scale confound hypothesis} ($\mathcal{H}_{\text{scale}}$): larger models may dominate both the CLB diagnostic and miniF2F accuracy simultaneously, with raw parameter count---rather than systematic compositional generalisation---as the true latent predictor. To formally adjudicate between $\mathcal{H}_{\text{CLB}}$ and $\mathcal{H}_{\text{scale}}$, 
we deploy a competitive hypothesis testing tournament. This is motivated by the non-nested model selection logic of ~\citet{vuong1989likelihood} but operating entirely within an exact permutation framework ~\citep{pitman1937significance,pesarin2010permutation}. 

\textbf{Choice of Test Statistic: Pearson $r$ Under Permutation.}
We adopt Pearson's product-moment correlation coefficient $r$ as the test statistic for the continuous boundary test. A common misconception holds that using $r$ forces bivariate normality assumptions; this is only true when $p$-values are derived from asymptotic Student's $t$-distributions. When embedded in an exact permutation framework, $r$ operates as a purely deterministic summary of directional coupling, with no distributional preconditions~\citep{pitman1937significance}. Relative to rank-based alternatives such as Spearman's $\varrho$, Pearson $r$ better captures proportional metric gradients: a model that breaks a structural bottleneck and scales sharply on the miniF2F axis in proportion to its CLB gain produces a stronger $r$ signal. Critically, by shifting the construction of the null distribution entirely to exact permutation enumeration, we bypass the standard requirements of homoscedasticity and normality that would otherwise render our small-sample analysis invalid~\citep{pesarin2010permutation}.

\textbf{Continuous Space Permutation Distributions.}
All analyses use $N=10$ consistently. Under $\mathcal{H}_0^{\text{continuous}}$, miniF2F performance is structurally independent of the predictor; any pairing of a model's predictor value with another model's miniF2F score is equally likely. We fix the miniF2F vector and compute $r$ across all $10! = 3{,}628{,}800$ permutations of each predictor, forming an exact empirical null distribution:
\begin{equation}
    p = \frac{1}{N!}\sum_{i=1}^{N!} \mathbb{I}\!\left(r_{\pi_i} \ge r_{\text{obs}}\right)
\end{equation}
For $\mathcal{H}_{\text{CLB}}$ (adj-ZSCT), the observed correlation $r = 0.7652$ yields an exact $p$-value of $\mathbf{p_{\text{CLB}}^{\text{continuous}}=0.00692}$---highly significant. For $\mathcal{H}_{\text{scale}}$ (total parameters in billions), the observed correlation $r = 0.3597$ yields $p_{\text{scale}}^{\text{continuous}} = 0.17837$, providing no evidence of a continuous dependency between parameter count and miniF2F accuracy.

\textbf{Scale Tail Partition Test (N=10, consistent).}
The most direct rebuttal to the scale hypothesis is \textit{DeepSeek-Prover-V2-7B}, which achieves $75.6\%$ miniF2F on only 7B parameters---placing it in the hard tail while remaining the smallest model in the suite.
This single counterexample directly falsifies scale as a \emph{necessary} condition for Olympiad-level performance.
We corroborate this with an exact tail partition test using total parameter count as the predictor ($N=10$ consistently).
The five hard-tail models yield an aggregate parameter sum of 790B.
Enumerating all $\binom{10}{5} = 252$ partitions, 6 meet or exceed this value ($p_{\text{scale}}^{\text{tail}} = 0.02381$).
By contrast, the CLB tail test yields exactly 1 out of 252 partitions ($p_{\text{CLB}}^{\text{tail}} = 0.00397$).

\textbf{Non-Parametric Competitive Verdict.}
Under the exact permutation framework~\citep{pitman1937significance,pesarin2010permutation}, CLB competency achieves significance on both the continuous ($p_{\text{CLB}}^{\text{continuous}} = 0.0069$) and partition ($p_{\text{CLB}}^{\text{tail}} = 0.0040$) axes.
Scale establishes no continuous footprint ($p_{\text{scale}}^{\text{continuous}} = 0.178$) and, while marginally significant in the tail partition ($p_{\text{scale}}^{\text{tail}} = 0.024$), is directly refuted as a necessary condition by the DP-V2-7B counterexample.
Systematic compositional generalisation is thus the dominant structural predictor of elite formal verification performance.

\begin{tcolorbox}[
  enhanced,
  colback=blue!6, colframe=blue!30, boxrule=0.6pt, arc=3pt,
  left=6pt, right=6pt, top=4pt, bottom=4pt,
]
\textbf{Takeaway 2.} CLB competency achieves significance on both the continuous ($p_{\text{CLB}}^{\text{continuous}} = 0.0069$) and partition ($p_{\text{CLB}}^{\text{tail}} = 0.0040$) axes.
Raw parameter count achieves no continuous signal ($p_{\text{scale}}^{\text{continuous}} = 0.178$) and is directly refuted as a necessary condition by DP-V2-7B (7B params, in the hard tail), ruling out scale as the driving confound.
\end{tcolorbox}

\section{Methodological Limitations}
\label{app:methodology}

Evaluating frontier foundational models introduces severe sample constraints: our pool ($N=10$) is strictly bounded by the immense compute requirements of training and running state-of-the-art provers. Operating within a small-sample regime limits the use of traditional parametric or asymptotic statistical tools, rendering standard regression or continuous correlation metrics structurally invalid.

\textbf{Inadequacy of Parametric and Asymptotic Estimators.}
Standard statistical methodologies depend on asymptotic behaviours, where normality and uniform variance are assumed as sample sizes approach infinity ($N \to \infty$). Applying these to an $N=10$ evaluation space introduces critical vulnerabilities:
\begin{itemize}
    \item \textbf{Parametric Inferences:} Testing linear regression models via Student's $t$ or $F$-statistics requires normally distributed residuals. In a 10-point dataset, testing for normality using standard diagnostics lacks sufficient statistical power, making significance claims highly sensitive to subtle outliers.
    \item \textbf{Quantile Regression:} While boundary limits are traditionally evaluated via high-quantile curves (e.g., tracking the 90th percentile), these estimators require dense data along the margins of the distribution. At $N=10$, estimating high-quantile paths lacks the necessary degrees of freedom, creating unstable, overfit boundaries that track individual peripheral data points.
    \item \textbf{Heteroscedastic Distortions:} Prerequisite capability boundaries inherently display non-uniform variance (heteroscedasticity). Low input capabilities restrict output variance tightly near a baseline floor, while high input scores unlock the full vertical range of execution. Standard Ordinary Least Squares (OLS) models underperform under these conditions, distorting standard errors and inflating the risk of false-positive significance claims.
\end{itemize}

\textbf{The Rigor of Exact Permutation Tools.}
To circumvent these small-sample limitations, our pipeline uses exact permutation tests processed via verified open-source scientific computing engines \cite{good2005permutation, virtanen2020scipy}. Permutation methods offer an elegant, distribution-free framework that provides exact mathematical precision on small sample pools:
\begin{enumerate}
    \item \textbf{Zero Distributional Dependencies:} Permutation techniques operate without making structural assumptions regarding the normality or variance profiles of the underlying data points. They bypass population parameter estimation by calculating significance directly from the empirical dataset's internal combinatorial structure.
    \item \textbf{Deterministic Execution Limits:} When total possible dataset configurations remain low, numerical engines can directly process the entire factorial set of allocations ($10! = 3{,}628{,}800$ for the global pairing test, and $\binom{10}{5}=252$ for the tail partition). By calculating outcomes across the entire permutation space, the resulting $p$-value represents an exact combinatorial probability under the null hypothesis rather than an asymptotic approximation.
\end{enumerate}
By shifting from continuous parametric approximations to exact partition structures, we ensure that our confirmation of the tail-dependent bottleneck remains robust, reproducible, and completely free from small-sample estimation bias.

\textbf{On the scaling of Few-shot CoT prompting.}
In our experiments, few-shot CoT prompting with $N=10$ exemplars is shown sufficient in eliciting the CLB competences of a wide range of LMs, from 7B to 671B.
However, in-context learning capabilities, and with it the efficiency of few-shot CoT prompting, are known to scale with LM size~\citep{brown2020language, wei2022emergent, dong2022survey} and to depend on model family.
Thus, it is not impossible that $N=10$ exemplars may simply not be enough for some of our tested models to perform as well as their competency allows.
However, we argue that this is fairly unlikely given the fact that the competency has been successfully elicited in LMs of size 7B, 8B, 32B, 72B, and 671B, spanning the whole range of tested model sizes and accounting for most of the different base model families represented in our evaluation (see Table~\ref{tab:zsct_minif2f}).

\end{document}